\pdfoutput=1

\documentclass[11pt]{article}

\usepackage{acl}

\usepackage{times}
\usepackage{latexsym}

\usepackage[T1]{fontenc}

\usepackage[utf8]{inputenc}

\usepackage{microtype}

\usepackage{multirow}
\usepackage{graphicx} 

\title{Using Two Losses and Two Datasets Simultaneously \\to Improve TempoWiC Accuracy}

\author{Mohammad Javad Pirhadi, Motahhare Mirzaei \and Sauleh Eetemadi \\
  Iran University of Science and Technology at Tehran, Iran \\
  \texttt{\{mohammad\_pirhadi, m\_mirzaei96\}@comp.iust.ac.ir},
  \texttt{sauleh@iust.ac.ir} \\}

\begin{document}
\maketitle
\begin{abstract}
WSD (Word Sense Disambiguation) is the task of identifying which sense of a word is meant in a sentence or other segment of text. Researchers have worked on this task (e.g. \citealp{Pustejovsky:02}) for years but it’s still a challenging one even for SOTA (state-of-the-art) LMs (language models). The new dataset, TempoWiC introduced by \citet{loureiro-etal-2022-tempowic} focuses on the fact that words change over time. Their best baseline achieves 70.33\% macro-F1. In this work, we use two different losses simultaneously to train RoBERTa-based classification models. We also improve our model by using another similar dataset to generalize better. Our best configuration beats their best baseline by 4.23\% and reaches 74.56\% macro-F1.
\end{abstract}

\section{Introduction}

In 2019, \citet{DBLP:journals/corr/abs-1808-09121} introduced WiC dataset. It is framed as a binary classification task between pairs of sentences including one identical target word with different meanings. In 2020, XL-WiC was introduced by \citet{DBLP:journals/corr/abs-2010-06478} and made WiC richer by providing more examples and adding more languages. We benefit from the English part of XL-WiC as a helping dataset to improve the generalization of our model. 

\citet{loureiro-etal-2022-tempowic} baselines include: RoBERTa \citep{DBLP:journals/corr/abs-1907-11692} base and large, TimeLMs \citep{loureiro-etal-2022-timelms} 2019-90M and 2021-124M and BERTweet \citep{nguyen-etal-2020-bertweet} base and large. They examine two different methods of using these models: Fine-tuning and SP-WSD layer pooling weights as explained in \citet{LOUREIRO2022103661}. The best result is for TimeLMs-2019-90M with SP-WSD with 70.33\% macro-F1. We examine RoBERTa-base and TimeLMs-Jun2022-153M.

For classification, many previous works (e.g. \citealp{DBLP:journals/corr/abs-1909-04164}) use standard practice and concatenate both sentences with a \texttt{[SEP]} token and fine-tune the \texttt{[CLS]} embedding. In this work, we use two different losses simultaneously, cross entropy loss on RoBERTa classification head output as standard practice and add cosine embedding loss on average of target word output embeddings. 

\section{Methodology}

\subsection{Model}

\begin{figure*}
    \centering
    \scalebox{0.525}{\includegraphics{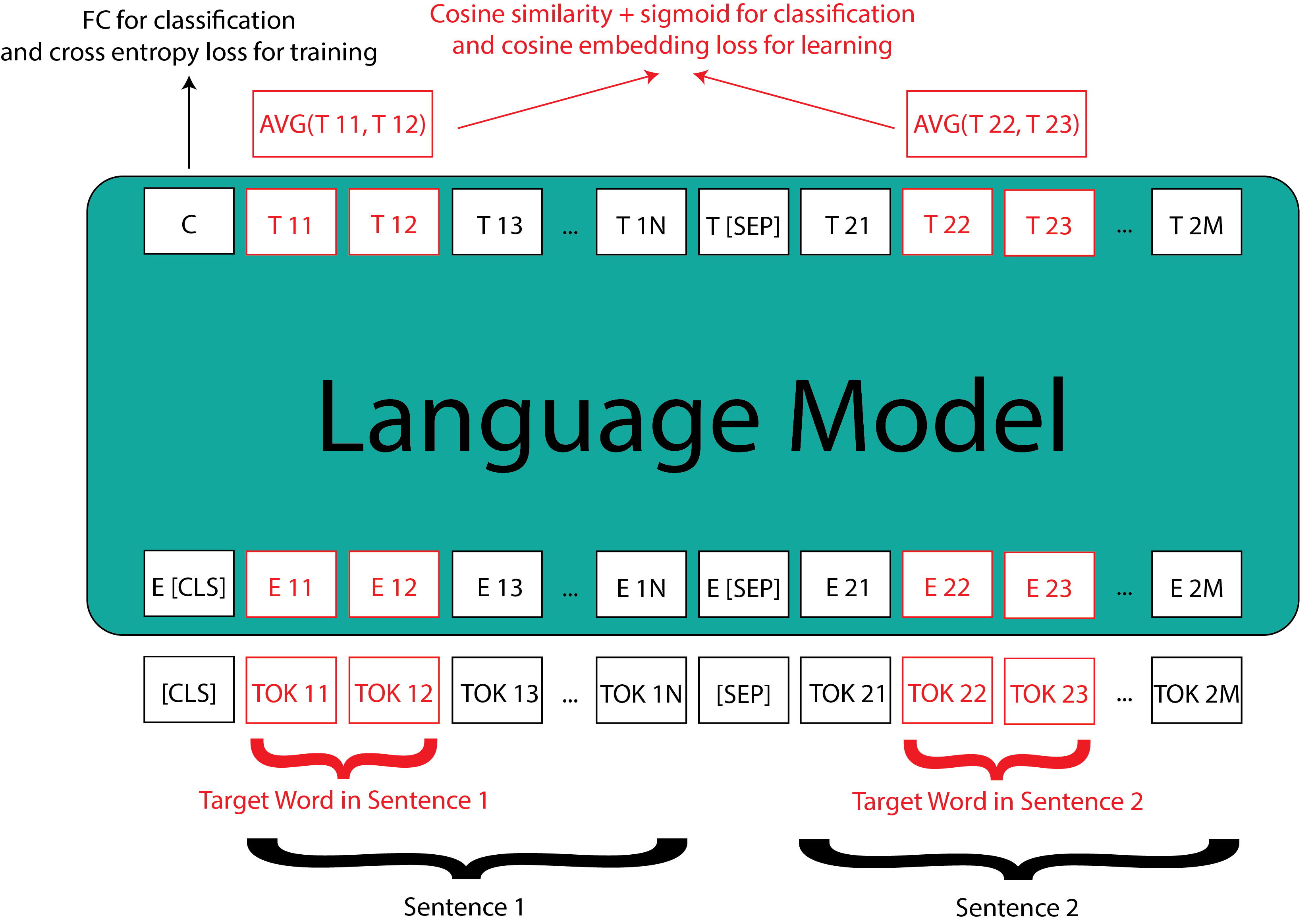}}
    \caption{An overview of architecture. We use two losses simultaneously. First one is cross entropy loss on standard classifier head (black path) and the second one is cosine embedding loss on average of target word output embeddings (red path).}
    \label{fig:architecture}
\end{figure*}

We use LMs as base model. We add classification head and also cosine similarity + sigmoid on top of them. The classification head consists of two FC (fully connected) layers and a dropout layer between them (like standard RoBERTa classification head). RoBERTa uses a byte-level BPE (Byte-Pair Encoding) encoding scheme so it's possible that we have multiple embeddings for a single word. For second output path, we average embeddings (it can be more than one as explained) related to target word in first sentence and second sentence and compare them using cosine similarity, finally we apply sigmoid activation to get a binary classification. Our experiments shows that the second output path is more accurate by a large margin. Figure \ref{fig:architecture} shows an overview of described architecture.

\subsection{Loss Function}

For the loss function, we have the sum of two losses, one on standard RoBERTa classification head and another on similarity (in case of the same meaning) or dissimilarity (in case of the different meaning) of the embeddings of the last layer related to the target word. For the former, we use cross entropy loss and for latter we use cosine embedding loss. The second loss, help our model to make similar contextual embeddings for target word closer and push dissimilar ones away from each other.

\subsection{Dataset}

The main dataset is TempoWiC, but we also use the XL-WiC dataset to make our model more robust. It's important to know that XL-WiC samples are not tweets so it is out-of-domain data and the added data may cause model accuracy to degrade if the combined dataset is not representative. We explored using the main dataset without adding any sample from the XL-WiC dataset, by adding a random subset of XL-WiC, and also by adding the whole XL-WiC.

\subsection{Framework \& Tools}

We use PyTorch \citep{NEURIPS2019_9015} + HuggingFace transformers \citep{wolf-etal-2020-transformers} to implement our models and for reporting results we use the Codalab online platform\footnote{\url{https://codalab.lisn.upsaclay.fr/competitions/5360}}.

\subsection{Hyper-parameters}

We use Ray Tune \citep{liaw2018tune} to tune our hyper-parameters including learning rate, train epochs, random seed, batch size and weight decay. Increasing weight decay helps us avoid over-fitting which was the main problem in our initial model.

\section{Experiments}

We have multiple configurations to test:
\begin{enumerate}
    \item Model
    \begin{itemize}
        \item RoBERTa-base
        \item TimeLMs-Jun2022-153M
    \end{itemize}
    \item Output
    \begin{itemize}
        \item Standard Classifier Head (FC)
        \item Cosine Similarity + Sigmoid (CS+S)
    \end{itemize}
    \item How we use XL-WiC
    \begin{itemize}
        \item Do not use (No)
        \item A subset as described (Sub)
        \item Whole XL-WiC (All)
    \end{itemize}
\end{enumerate}

\begin{table*}
\centering
\begin{tabular}{cccc}
\hline
\textbf{Output} &  \textbf{Model} & \textbf{XL-WiC Use} &\textbf{Macro-F1}\\
\hline
\multirow{6}{*}{Classifier} & \multirow{3}{*}{RoBERTa-base} & No & 62.35\% \\\cline{3-4}
 & & Sub & 65.98\% \\\cline{3-4}
 & & All & 63.56\%\\\cline{2-4}
 & \multirow{3}{*}{TimeLMs-Jun2022-153M} & No & 64.54\% \\\cline{3-4}
 & & Sub & 73.16\% \\\cline{3-4}
 & & All & 72.77\% \\
 \hline
\multirow{6}{*}{Similarity} & \multirow{3}{*}{RoBERTa-base} & No & 	67.26\% \\\cline{3-4}
 & & Sub & 68.29\% \\\cline{3-4}
 & & All & 67.29\% \\\cline{2-4}
 & \multirow{3}{*}{TimeLMs-Jun2022-153M} & No & 66.69\% \\\cline{3-4}
 & & Sub & \textbf{74.56\%} \\\cline{3-4}
 & & All & 72.32\% \\
\hline
\multirow{6}{*}{Official Baselines} & TimeLMs-2019-90M-SIM & No & 70.33\% \\\cline{2-4}
& RoBERTa-L-SIM & No & 67.09\% \\\cline{2-4}
& RoBERTa-L-FT & No & 59.10\% \\\cline{2-4}
& TimeLMs-2019-90M-FT & No & 57.70\% \\\cline{2-4}
& Random & No & 50.00\% \\\cline{2-4}
& All True & No & 26.79\% \\
\hline
\end{tabular}
\caption{\label{results}
All results are obtained from the Codalab online platform on Tempo-WiC test set.
}
\end{table*}

\subsection{Results}

The biggest problem we were facing was over-fitting. This is expected since we use transformer-based LMs.

The most accurate configuration is TimeLMs-Jun2022-153M with cosine similarity + sigmoid output trained on TempoWiC and a subset of XL-WiC. In the following paragraphs, we are going to analysis the results.

First, the results show that TimeLMs-Jun2022-153M beats RoBERTa in all possible configurations, the reason is simple: TempoWiC consists of tweets and TimeLMs-Jun2022-153M is trained on tweets too, but RoBERTa is not trained on tweets.

Second, using XL-WiC improves results in all cases. Using all XL-WiC example reduces the accuracy because the distribution is different (the samples are not tweets) and it has almost $4\times$ data in comparison to TempoWiC. If we use all of its data, we can not expect better accuracy because the training set distribution will be different from the test distribution. 

Last, the cosine similarity + sigmoid output is better in most cases in comparison to the standard classifier head. We think it's because of more focus on the target word embedding in comparison to more focus on the whole context.

\section{Future Work}

In the future work, more configurations can be explored:
\begin{enumerate}
    \item Selecting the subset of XL-WiC more wisely, instead of randomly selecting. For example, considering the maximum possible use of unique words.
    \item Using more layers to calculate similarity, instead of using only the last layer. For example, the sum of the last 4 layers is another common choice in word sense disambiguation settings.
    \item Exploring more similarity functions, instead of cosine similarity. For example, euclidean distance can also be explored.
\end{enumerate}

\section{Conclusion}

In this work, we beat the best baseline of \citet{loureiro-etal-2022-tempowic} by a large margin. To do this we use two losses simultaneously (standard classifier head cross entropy loss and cosine embedding loss on average of target word output embeddings) to train SOTA LMs, and also use XL-WiC as a helping dataset to generalize better. The best LM was TimeLMs-Jun2022-153M which is a pre-trained model on 153M tweets.

\section{Acknowledgements}
We would like to express our special thanks of gratitude to Mohammad Mahdi Javid who helped us with preparing training resources.

\bibliography{acl}
\bibliographystyle{acl_natbib}

\end{document}